\documentclass[pdflatex,sn-mathphys-num]{sn-jnl}

\usepackage{graphicx}%
\usepackage{multirow}%
\usepackage{amsmath,amssymb,amsfonts}%
\usepackage{amsthm}%
\usepackage{mathrsfs}%
\usepackage[title]{appendix}%
\usepackage{xcolor}%
\usepackage{textcomp}%
\usepackage{manyfoot}%
\usepackage{booktabs}%
\usepackage{algorithm}%
\usepackage{algorithmicx}%
\usepackage{algpseudocode}%
\usepackage{listings}%
\usepackage{pgfplots}
\usepackage{pgfplotstable}
\usepackage{filecontents}
\usepackage{caption}
\usepackage{hyperref}
\pgfplotsset{compat=1.17} 
\usepackage{filecontents} 

\theoremstyle{thmstyleone}%
\theoremstyle{thmstyletwo}%

\theoremstyle{thmstylethree}%

\begin{document}

\title[Article Title]{AlgoPilot: Fully Autonomous Program Synthesis Without Human-Written Programs}


\author*[1]{\fnm{Xiaoxin} \sur{Yin}}\email{xiaoxin@gmail.com}

\affil*[1]{\orgdiv{Independent Researcher}}


\abstract{Program synthesis has traditionally relied on human-provided specifications, examples, or prior knowledge to generate functional algorithms. Existing methods either emulate human-written algorithms or solve specific tasks without generating reusable programmatic logic, limiting their ability to create novel algorithms. We introduce AlgoPilot, a groundbreaking approach for fully automated program synthesis without human-written programs or trajectories. AlgoPilot leverages reinforcement learning (RL) guided by a Trajectory Language Model (TLM) to synthesize algorithms from scratch. The TLM, trained on trajectories generated by random Python functions, serves as a soft constraint during the RL process, aligning generated sequences with patterns likely to represent valid algorithms. Using sorting as a test case, AlgoPilot demonstrates its ability to generate trajectories that are interpretable as classical algorithms, such as Bubble Sort, while operating without prior algorithmic knowledge. This work establishes a new paradigm for algorithm discovery and lays the groundwork for future advancements in autonomous program synthesis.}




\maketitle

\section{Introduction}\label{sec1}

Program synthesis is the task of automatically generating programs that satisfy a given specification, typically expressed as input-output examples, formal constraints, or high-level descriptions. The ultimate goal of program synthesis is to bridge the gap between human intent and executable code, enabling systems that can generate reliable and correct programs with minimal human intervention. The development of deep learning has significantly advanced program synthesis by enabling models to learn complex program structures and semantics directly from data. Sequence-to-sequence models and transformers have been successfully applied to generate programs from natural language descriptions, input-output examples, or partial specifications~\cite{devlin2017robustfill, balog2017deepcoder}. For instance, \textit{DeepCoder} by Balog et al.~\cite{balog2017deepcoder} demonstrated the potential of neural-guided search by predicting program sketches and efficiently guiding symbolic solvers. Similarly, \textit{RobustFill} by Devlin et al.~\cite{devlin2017robustfill} employed sequence-to-sequence models to synthesize string transformation programs from input-output pairs, showcasing the capabilities of deep neural networks in capturing program patterns.

In recent years there are two main directions of using deep learning for program synthesis:

\begin{itemize}
    \item \emph{Neural Reasoning Models}, which learn to program with training data being the ``trajectories'' of human-written algorithms. 
    
      \par Neural reasoning models aim at emulating algorithmic and programmatic reasoning through end-to-end learning paradigms. Approaches such as Neural Programmer-Interpreters (NPI) \cite{reed2015neural}, A Generalist Neural Algorithmic Learner (GNAL) \cite{velivckovic2021neural}, and Neural Algorithmic Reasoning with Causal Regularisation \cite{zhang2022causal} share a common methodology: they train neural models using trajectories derived from intermediate states of human-written programs. 
      
      \par However, these approaches can only rebuild a program based on a human-written program. They cannot create algorithms or programs on their own (e.g., based on inputs and desired outputs). Thus their practical application is limited.

    \item \emph{End-to-End Neural Task Solvers}, which train models to directly solve tasks (such as sorting) instead of creating programs.

      \par Another line of research focuses on training models to complete tasks typically performed by classical algorithms, such as sorting. Notable examples include Neural Turing Machines (NTMs) \cite{graves2014neural}, Differentiable Neural Computers (DNCs) \cite{graves2016hybrid}, and AlphaDev \cite{alphadev2023}. Unlike traditional approaches that construct algorithms using explicit operations like "Compare" and "Swap," these models operate differently. They are trained end-to-end and can observe all inputs simultaneously, enabling them to learn task-specific behaviors directly.

      \par However, this is very different from what we mean by ``algorithms''. For example, a sorting algorithm should only use ``Compare'' and ``Swap'' (or ``Copy'') operations, instead of using a DNN to check all inputs at the same time and select the smallest element. The above approaches cannot create algorithms that can be implemented by a typical programming language.

\end{itemize}

We can see that neither of the two lines of research can automatically create a program to solve an algorithmic task on its own. \emph{Neural Reasoning Models} rely on human-created programs to provide training data. \emph{End-to-End Neural Task Solvers} create models instead of programs. Here comes our key question: \textbf{Is it possible to train a model that can create an algorithm on its own?}

An early work along this direction is presented by Abolafia et al. \cite{abolafia2018neural}, who used RNNs to synthesize programs in BF, a very simple programming language. It only requires the inputs and desired outputs of the program, without any human-written programs. However, it can only solve very simple problems such as arithmetic operations and basic string manipulations, and is not powerful enough to solve algorithmic tasks such as sorting. 

In this paper we present \emph{AlgoPilot}, the first approach that can learn to create an algorithm purely on its own, without human-created algorithms or trajectories. \emph{AlgoPilot} uses reinforcement learning to learn to accomplish a algorithmic task, such as sorting. Comparing with existing methods that learns from trajectories of human-written algorithms \cite{reed2015neural} \cite{velivckovic2021neural} \cite{zhang2022causal}, \emph{AlgoPilot} uses a language model to build a model for predicting the next token in a trajectory of a randomly generated python function. This language model's prediction becomes a soft constraint for its reinforcement learning (a.k.a. RL) process, which guides the RL model to generate trajectories that can likely be generated by a real algorithm. This enables \emph{AlgoPilot} to train a model which generates trajectories that are highly similar to those of a real algorithm. One can easily find an algorithm (or a python function) that generates such trajectories (possibly with the help of LLMs), and in this way \emph{AlgoPilot} achieves automated program synthesis without human instructions.

Let us take sorting with double-loop as an example. Here is the procedure of \emph{AlgoPilot} in generating a sorting algorithm with a double-loop:

\begin{enumerate}
    \item \textbf{Random Function Generator}
    
        \par \noindent We create a random function generator that can generate a random python function with a double-loop, and randomly add the two operations ``\emph{Compare}'' and ``\emph{Swap}''.
        
    \item \textbf{Trajectory Language Model (TLM)}
    
        \par \noindent Many random functions are generated, each producing a trajectory of relevant operations (\emph{Compare}, \emph{Swap}, and the indices involved). \emph{AlgoPilot} trains a language model of the trajectories, which can predict the probability of observing a token after a prefix of tokens.
        
    \item \textbf{Reinforcement Learning}
    
        \par \noindent \emph{AlgoPilot} uses reinforcement learning to train a transformer model for sorting. Given the input's length, the model can only use \emph{Compare} and \emph{Swap} operations, and the environment only returns one feedback for each operation. For ``\emph{Compare i j}'', the environment returns if the element at position \emph{i} is smaller than, equal to, or greater than that at position \emph{j}. For \emph{Swap}, the environment returns a boolean indicating whether the array is sorted. Rewards are given to the model when it finds a pair of elements that need to be swapped, or successfully sorts the array.
        
    \item \textbf{Guided Reinforcement Learning}
    
        \par \noindent The above reinforcement learning can train a model that can sort an array. However, its trajectory does not follow any obvious pattern, and it does not seem feasible to create an algorithm based on the trajectory. Therefore, \emph{AlgoPilot} uses the \emph{Trajectory Language Model (TLM)} to enhance the reinforcement learning process. It adds an additional reward when the next operation has a high probability according to the \emph{TLM}, and adds a penalty for a low probability. This guides the model to learn to generate a trajectory that can be produced by an algorithm.

    \item \textbf{Algorithm Creation}

        \par \noindent The trajectory generated in Step 4 usually follows obvious algorithmic patterns. Currently \emph{AlgoPilot} uses LLMs (such as GPT-4o) to generate an algorithm or a python function. But we can certain train a language model to generate them from the trajectories of randomly generated functions.
    
\end{enumerate}

The rest of the paper is organized as follows. Related work is reviewed in Section 2. Section 3 presents \emph{AlgoPilot}, including the reinforcement learning environment, random function generator, trajectory language model, guided reinforcement learning, and finally the creation of algorithm. Experiment results are included in each subsection. We then discuss about the future work in Section 4, and conclude this paper in Section 5.

\section{Related Work}\label{sec2}

Historically, program synthesis can be traced back to early theoretical work in the 1960s, such as Church's synthesis problem, which posed the challenge of finding a program satisfying a logical specification. Since then, the field has evolved significantly, driven by advances in computational power, formal verification techniques, and learning-based methods.

In the early 2000s, program synthesis gained momentum with the introduction of inductive program synthesis approaches, where programs are synthesized from input-output examples. Notable among these is the \textit{Programming by Example (PBE)} paradigm, which allows users to specify program behavior through examples rather than formal specifications. Tools such as \textit{FlashFill} by Gulwani et al.~\cite{gulwani2011automating} demonstrated the practical feasibility of program synthesis in spreadsheet automation, enabling non-programmers to perform complex tasks via example-driven specifications. Around the same time, \textit{SKETCH}~\cite{solar2008program} introduced a counterexample-guided inductive synthesis (CEGIS) approach, which iteratively refines candidate programs based on counterexamples generated by a verifier. Additionally, \textit{Angelic Programming}~\cite{bodik2007angelic} explored angelic non-determinism to simplify program search by focusing on high-level intent. These pioneering efforts laid the foundation for modern approaches to program synthesis, setting the stage for incorporating machine learning, neural networks, and reinforcement learning techniques in subsequent years.

In recent years, deep learning has significantly advanced program synthesis by enabling models to learn complex program structures and semantics directly from data. Traditional search-based and symbolic methods often struggle with large search spaces, while neural approaches offer a more scalable and flexible alternative. Deep learning models, particularly sequence-to-sequence architectures and transformers, have been successfully applied to generate programs from natural language descriptions, input-output examples, or partial specifications~\cite{devlin2017robustfill, balog2017deepcoder}. For instance, \textit{DeepCoder} by Balog et al.~\cite{balog2017deepcoder} demonstrated the potential of neural-guided search by predicting program sketches and efficiently guiding symbolic solvers. Similarly, \textit{RobustFill} by Devlin et al.~\cite{devlin2017robustfill} employed sequence-to-sequence models to synthesize string transformation programs from input-output pairs, showcasing the capabilities of deep neural networks in capturing program patterns. The advent of large-scale pre-trained language models, such as OpenAI's Codex~\cite{chen2021evaluating}, has further pushed the boundaries by generating executable code from natural language prompts. These advancements illustrate the growing synergy between deep learning techniques and program synthesis, opening up new possibilities for automating software development and improving programming productivity.

Abolafia et al. \cite{abolafia2018neural} introduced an approach leveraging recurrent neural networks (RNNs) to synthesize programs in BF, a very simple programming language. Their method employs Priority Queue Training (PQT), an iterative optimization strategy that maintains a queue of top-performing programs, focusing the training on promising samples. This approach addresses the challenge of sparse rewards in program synthesis and demonstrates improved generalization across algorithmic tasks, including string manipulation and arithmetic operations. The results highlight the potential of neural networks in capturing logical reasoning patterns essential for end-to-end program synthesis.

\section{\emph{AlgoPilot}: Automated Learning of Algorithm without Human Help}

In this section we present \emph{AlgoPilot}, the first approach that can learn to create an algorithm purely on its own, without human-created algorithms or trajectories. \emph{AlgoPilot} uses reinforcement learning to learn to accomplish a algorithmic task. In this study we use \emph{sorting with double-loop} as the example, and the approach can be applied to many other types of algorithms.

Comparing with existing methods that learns from trajectories of human-written algorithms \cite{reed2015neural} \cite{velivckovic2021neural} \cite{zhang2022causal}, we generate many random python functions with double-loops, and train a language model to predict the next token in a trajectory of a randomly generated python function. The prediction of this language model becomes a soft constraint for its reinforcement learning process, which guides the RL model to generate trajectories that can likely be generated by a real algorithm. This enables \emph{AlgoPilot} to train a model which generates trajectories that are highly similar to those of a real algorithm.

\subsection{Learning to Sort with Reinforcement Learning} \label{ssec:rl}

\subsubsection{Environment of Reinforcement Learning} 

Before going into the details of \emph{AlgoPilot}, we first explain our setting of learning to sort with reinforcement learning. We use a \emph{Sorting Environment} to manage the state and dynamics of the sorting task. Each episode begins with generation of a random array of integers with a specified length (6 to 14).  The agent interacts with the environment through a series of actions, which includes actions like ``\emph{Compare}'' and ``\emph{Swap}'', along with tokens representing various comparison outcomes and list indices.

When the agent performs a \emph{Compare} action, it selects two indices from the list to compare their values, receiving feedback on whether the first is less than, equal to, or greater than the second. If the agent chooses to \emph{Swap} two elements, the environment swaps them in the underlying array, and gives a reward or penalty based on whether the two elements are swapped into the right order. The environment also imposes a maximum number of steps per episode to encourage efficiency and prevent endless interactions.

Reward mechanisms are designed to guide the agent toward effective sorting. Positive rewards are given for successful swaps and achieving a sorted list, while negative rewards penalize invalid actions. A small negative reward is given to every step, in order to penalize inefficient moves. The environment employs a combination of long-term gamma (for finally achieving a sorted array) and short-term gamma (for all other rewards) in the Bellman equations, in order to balance immediate and long-term rewards, fostering strategies that are both effective and efficient over the course of an episode.

\subsubsection{Reinforcement Learning Agent} 

We use a simple transformer model for this reinforcement learning task, which predicts actions within the sorting environment. Unless otherwise specified, our default transformer has 4 layers, 8 heads, and a hidden dimension of 192. Sinusoidal positional vectors are used for the position embeddings. We use the \texttt{TransformerEncoderLayer} of Pytorch, and Relu activation function.

Our reinforcement learning model is very different from many existing approaches such as Neural Turing Machines (NTMs) \cite{graves2014neural}, Differentiable Neural Computers (DNCs) \cite{graves2016hybrid}, and AlphaDev \cite{alphadev2023}. The above models observe the whole input array simultaneously, which enables them to directly select the smallest element or sorting the array. In contrast, our transformer-based agent can only use ``\emph{Compare}'' operations to learn about the underlying array, which is consistent with most sorting algorithms including Selection Sort, Bubble Sort, QuickSort and Merge Sort.

Please note that although our agent can learn to produce a sequence of ``\emph{Compare}'' and ``\emph{Swap}'' actions to sort an array, such sequences seldom follow any algorithmic patterns, and it is difficult to create an algorithm based on such sequences.

\subsubsection{Experiment Results}

Before going into more details of \emph{AlgoPilot}, we first present our experiment results of our simple reinforcement learning agent in this sorting environment. All our experiments were done using a computer with a NVIDIA A6000 GPU, an Intel i7-12700K CPU, with Ubuntu 18.04 and Pytorch 2.0.0. 

The sorting environment is configured as follows. The reward of sorting success is 0.5, which uses a gamma of 0.99 (i.e., it barely decays over the steps). All other rewards have a gamma of 0.7, which decays quickly over several steps. Each step has a reward of -0.3, and each swap that moves a smaller element to the front has a reward of 1.0. 

We study random arrays of different sizes (from 6 to 14, step 2). For each array size, we allow the model to perform at most $3 \cdot array\_size^2 $ operations. If the array is sorted before this limit on \#operations is hit, we say the model successfully sorts this array. 

We use an Epsilon-greedy policy for exploration-exploitation trade-off. Epsilon is set to 0.5 at the beginning, and will gradually drop to 0.05, which decays by a factor of $e$ every 1000 episodes. We run 50,000 episodes in each test. Learning rate is set to $1e-4$.

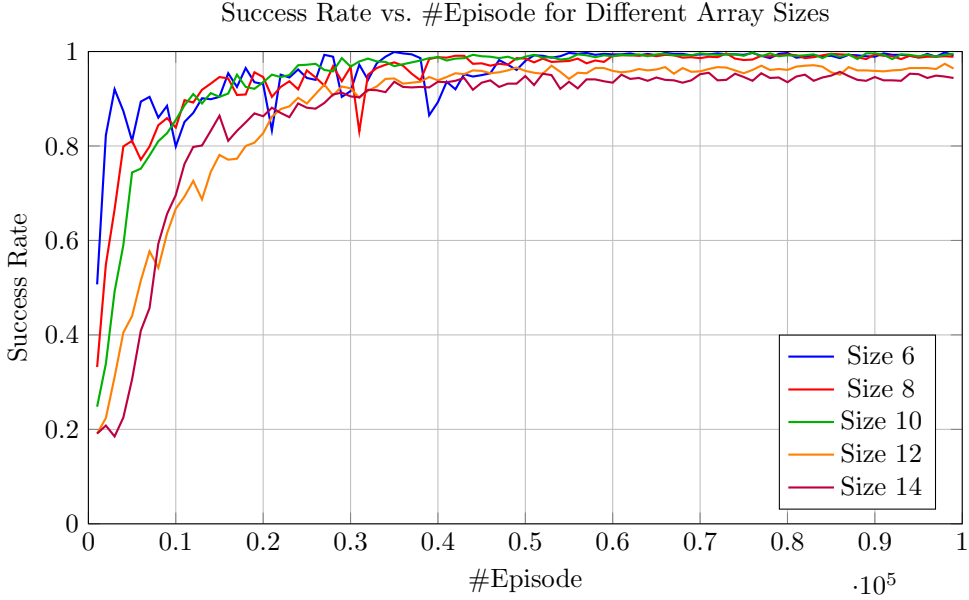
\begin{figure}[htbp]
  \centering
  \begin{tikzpicture}
    \begin{axis}[
        width=\textwidth,
        height=0.6\textwidth,
        xlabel={\#Episode},
        ylabel={Success Rate},
        title={Success Rate vs. \#Episode for Different Array Sizes},
        legend pos=south east,
        grid=both,
        xmin=0, xmax=100000,
        ymin=0, ymax=1,
        legend columns=1,
    ]
      \addplot[blue, thick] table[x=Episodes, y=Size6, col sep=comma] {data.csv};
      \addlegendentry{Size 6}
      
      \addplot[red, thick] table[x=Episodes, y=Size8, col sep=comma] {data.csv};
      \addlegendentry{Size 8}
      
      \addplot[green!70!black, thick] table[x=Episodes, y=Size10, col sep=comma] {data.csv};
      \addlegendentry{Size 10}
      
      \addplot[orange, thick] table[x=Episodes, y=Size12, col sep=comma] {data.csv};
      \addlegendentry{Size 12}
      
      \addplot[purple, thick] table[x=Episodes, y=Size14, col sep=comma] {data.csv};
      \addlegendentry{Size 14}
    \end{axis}
  \end{tikzpicture}
  \caption{Success Rate vs. \#Episode for various array sizes. We run 100,000 \#Episode for each array size from 6 to 14 (step 2), and have at least about 95\% success rate in each setting, which means our model successfully sort the array within $3 \cdot array\_size^2 $ operations.}
  \label{fig:success_rate}
\end{figure}

We focus on two metrics for our reinforcement learning model: \textbf{Success rate} and \textbf{Number of operations} (both \emph{Compare} and \emph{Swap}). Figure \ref{fig:success_rate} shows the success rate vs. number of episodes, which illustrates how the model learns to sort an array during the training process of reinforcement learning. We can see the success rate is close to 100\% for array sizes 6, 8 and 10. The success rate is around 95\% for array sizes 12 and 14\footnote{We tried array size of 16 and the model did not achieve a significant success rate}.


Now let us see how our model compares with QuickSort, a close-to-optimal sorting algorithm using only  ``\emph{Compare}'' and ``\emph{Swap}'' operations, in term of number of operations. The expected numbers of ``\emph{Compare}'' and ``\emph{Swap}'' operations of QuickSort are shown below, with details in Appendix \ref{secA1}.

\[
\begin{aligned}
\mathbb{E}[C(n)] &= 2\,(n+1)\,H_{n} - 2\,n, \\[6pt]
\mathbb{E}[S(n)] &= (n+1)\,H_{n} - 2\,n,
\end{aligned}
\]
where
\[
H_{n} = \sum_{k=1}^{n} \frac{1}{k}.
\]

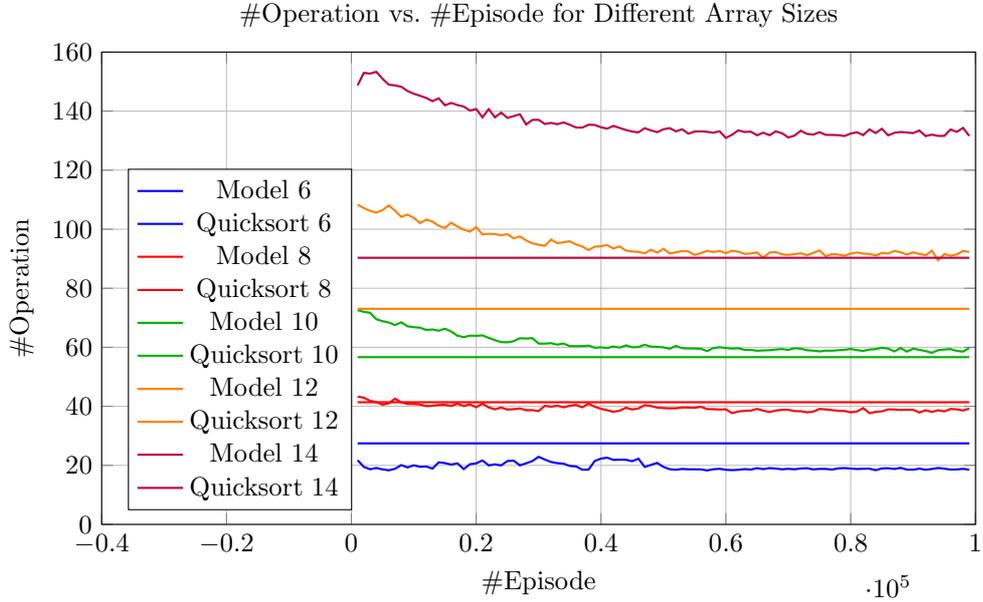
\begin{figure}[htbp]
  \centering
  \begin{tikzpicture}
    \begin{axis}[
        width=\textwidth,
        height=0.6\textwidth,
        xlabel={\#Episode},
        ylabel={\#Operation},
        title={\#Operation vs. \#Episode for Different Array Sizes},
        legend pos=south west,
        grid=both,
        xmin=-40000, xmax=100000,
        ymin=0, ymax=160,
        legend columns=1,
    ]
      \addplot[blue, thick] table[x=Episodes, y=Model6, col sep=comma] {operations.csv};
      \addlegendentry{Model 6}

      \addplot[blue, thick] table[x=Episodes, y=Quicksort6, col sep=comma] {operations.csv};
      \addlegendentry{Quicksort 6}
      
      \addplot[red, thick] table[x=Episodes, y=Model8, col sep=comma] {operations.csv};
      \addlegendentry{Model 8}

      \addplot[red, thick] table[x=Episodes, y=Quicksort8, col sep=comma] {operations.csv};
      \addlegendentry{Quicksort 8}
      
      \addplot[green!70!black, thick] table[x=Episodes, y=Model10, col sep=comma] {operations.csv};
      \addlegendentry{Model 10}

      \addplot[green!70!black, thick] table[x=Episodes, y=Quicksort10, col sep=comma] {operations.csv};
      \addlegendentry{Quicksort 10}
      
      \addplot[orange, thick] table[x=Episodes, y=Model12, col sep=comma] {operations.csv};
      \addlegendentry{Model 12}

      \addplot[orange, thick] table[x=Episodes, y=Quicksort12, col sep=comma] {operations.csv};
      \addlegendentry{Quicksort 12}
      
      \addplot[purple, thick] table[x=Episodes, y=Model14, col sep=comma] {operations.csv};
      \addlegendentry{Model 14}

      \addplot[purple, thick] table[x=Episodes, y=Quicksort14, col sep=comma] {operations.csv};
      \addlegendentry{Quicksort 14}
    \end{axis}
  \end{tikzpicture}
  \caption{\#Operations vs. \#Episode for various array sizes. We run 100,000 \#Episode for each array size from 6 to 14 (step 2). Each horizontal line represents the expected number of operations of Quicksort for each array size, and the plot in the same color represents the number of operations of our model. We can see that the model uses fewer operations for smaller array sizes, and more operations than Quicksort for larger array sizes.}
  \label{fig:operations}
\end{figure}

Figure \ref{fig:operations} shows the number of operations used to successfully sort an array of each size. Each horizontal line represents the expected number of operations of Quicksort for each array size, and the plot in the same color represents the number of operations of our model. We can see that for each array size, the number of operations keeps getting lower in the training process (i.e., as \#Episodes grows). The model uses fewer operations for smaller array sizes, and more operations than Quicksort for larger array sizes. It seems that our model can learn a procedure that is quite optimal for smaller arrays. While for larger arrays the model's procedure is less optimal.


Although the model can learn to sort an array efficiently and effectively, its sequence of operations does not seem to follow any algorithmic pattern. Here is an example in which the model successfully sorts an array of size 6. The operations seems rather arbitrary and we cannot identify any pattern. 

\begin{quote}
    \texttt{len6 Compare 0 4 more Swap Compare 2 3 more Swap Compare 4 5 more Swap Compare 1 2 more Swap Compare 3 4 more Swap Compare 2 3 less Compare 0 1 less Compare 3 4 less Compare 4 5 more Swap}
\end{quote}

In the following subsections we will describe \emph{AlgoPilot}, which guides the model to learn to generate sequences of operations that follow algorithmic patterns.

\subsection{Random Function Generator} \label{ssec:randfunc}

In order to train a language model for trajectories (which are sequences of operations in our study) of programs, we need to create a random function generator which generates random functions of a particular type. In this study we focus on functions with double-loops. With help from ChatGPT-o1, we create a random function generator that can generate a python function with a double-loop, as shown in Algorithm 1 and Algorithm 2.

\begin{algorithm}
\caption{GENERATE\_RANDOM\_CONDITION}
\label{alg:generate_random_condition}
\begin{algorithmic}
\State \textbf{Input:} None
\State \textbf{Output:} A single line containing a random \texttt{if} statement using \texttt{Compare(...)}.

\State possible\_refs $\gets$ [``arr[i]'', ``arr[i-1]'', ``arr[i+1]'', ``arr[j]'', ``arr[j-1]'', ``arr[j+1]'', ``y'']
\State left\_expr $\gets$ \texttt{RandomChoice}(possible\_refs)
\State right\_expr $\gets$ \texttt{RandomChoice}(possible\_refs)
\State Ensure left\_expr $\neq$ right\_expr if possible

\State possible\_return\_checks $\gets$ [``$> 0$'', ``$< 0$'', ``$== 0$'', ``$>= 0$'', ``$<= 0$''] 
\State op\_to\_zero $\gets$ \texttt{RandomChoice}(possible\_return\_checks)

\State \textbf{return} \texttt{``if self.Compare(left\_expr, right\_expr) op\_to\_zero:''}
\end{algorithmic}
\end{algorithm}

\begin{algorithm}
\caption{GENERATE\_RANDOM\_FUNCTION}
\label{alg:generate_random_function}
\begin{algorithmic}
\State \textbf{Output:} A string defining a new method with nested loops and a random condition.

\State func\_name $\gets$ ``random\_func\_doubleloop'' 
\State func\_signature $\gets$ ``def random\_func\_doubleloop(self):''

\State arr $\gets$ self.arr \Comment{Reference the instance array}
\State n $\gets$ len(arr)

\State \textbf{Outer Loop:}
\State Randomly choose:
\State \hspace{1em} \texttt{for i in range(n - 1)}
\State \hspace{1em} \texttt{for i in range(n)}

\State With some probability:
\State \hspace{1em} y $\gets$ arr[i] 

\State \textbf{Inner Loop:}
\State Randomly choose:
\State \hspace{1em} \texttt{for j in range(i+1, n)}
\State \hspace{1em} \texttt{for j in range(n - i - 1)}

\State \textbf{Condition Inside Inner Loop:}
\State condition\_line $\gets$ \texttt{GENERATE\_RANDOM\_CONDITION()}
\State \texttt{if self.Compare(...):} 

\State \hspace{1em} Call \texttt{self.Swap()}

\State \textbf{return} The entire function definition as a string.
\end{algorithmic}
\end{algorithm}

Listing 1 shows an example random function generated, and it definitely cannot sort an array. In order to generate a trajectory, this function randomly outputs \emph{Compare} and \emph{Swap} operations. Please note that \emph{AlgoPilot} is not aware of what a random function can do. It only uses trajectories of such randomly generated functions to train a language model for trajectories, and never knows whether these trajectories lead to sorted array or not. The learning to sort is done using the reinforcement learning environment by rewarding useful operations (described in Section \ref{ssec:rl}) , which is completely independent from the generation of random functions. 

Here is an example trajectory: 

\begin{quote}
\texttt{len14 Compare 0 1 more Compare 1 1 less Swap Compare 2 1 more Compare 3 1 more Compare 4 1 more Compare 5 1 more Compare 6 1 more Compare 7 1 more Compare 8 1 more Compare 9 1 more Compare 10 1 more Compare 11 1 more Compare 12 1 more Compare 13 1 more Compare 1 2 less Compare 2 2 less Swap Compare 3 2 more \ldots}
\end{quote}

\begin{figure}[H]
    \centering
    \begin{lstlisting}[language=Python, caption={A randomly generated function with a double-loop}]
class RandomDoubleLoopFunction:

    def __init__(self, n=14):
        self.arr = list(range(1, n + 1))
        random.shuffle(self.arr)
        self.orig_arr = list(self.arr)
        self.idx1, self.idx2 = -1, -1
        self.output = ['len16']

    def Swap(self):
        i, j = self.idx1, self.idx2
        self.arr[i], self.arr[j] = self.arr[j], self.arr[i]
        self.output.append("Swap")

    def Compare(self, a, b):
        self.idx1, self.idx2 = self.orig_arr.index(a), self.orig_arr.index(b)
        outcome = "equal"
        if a > b: 
            outcome = "more"
        else: 
            outcome = "less"
        self.output.append(f"Compare {self.idx1} {self.idx2} {outcome}")
        return (a > b) - (a < b)

    def random_func_doubleloop(self):
        arr = self.arr
        n = len(arr)
        for i in range(n):
            y = arr[i]
            for j in range(i + 1, n):
                if self.Compare(arr[j-1], arr[i+1]) == 0:
                    self.Swap()
    \end{lstlisting}
\end{figure}

\subsection{Trajectory Language Model (TLM)} \label{ssec:tlm}

Given a sequence like ``\texttt{len16 Compare 0 1 more Compare 1 1 less Swap Compare 2 1 more Swap Compare 3 1 more \ldots }'', we want to train a language model to predict the probability of seeing each token given the prefix. Since the vocabulary is very small (with 36 different tokens), we prefer a medium-sized language model. We choose QWen-0.5B \cite{yang2024qwen25}, which has the best benchmark results among all models with $<1\text{B}$ parameters on the Hugging Face Open LLM Leaderboard\footnote{\url{https://huggingface.co/spaces/open-llm-leaderboard/open_llm_leaderboard}}. 

All our experiments were done using a computer with a NVIDIA A6000 GPU, an Intel i7-12700K CPU, with Ubuntu 18.04 and Pytorch 2.0.0. We created 1.5M randomly generated functions with double-loop, and generated a trajectory using each of them. These 1.5M trajectories were randomly split into a training set (1.44M) and a testing set (0.06M).


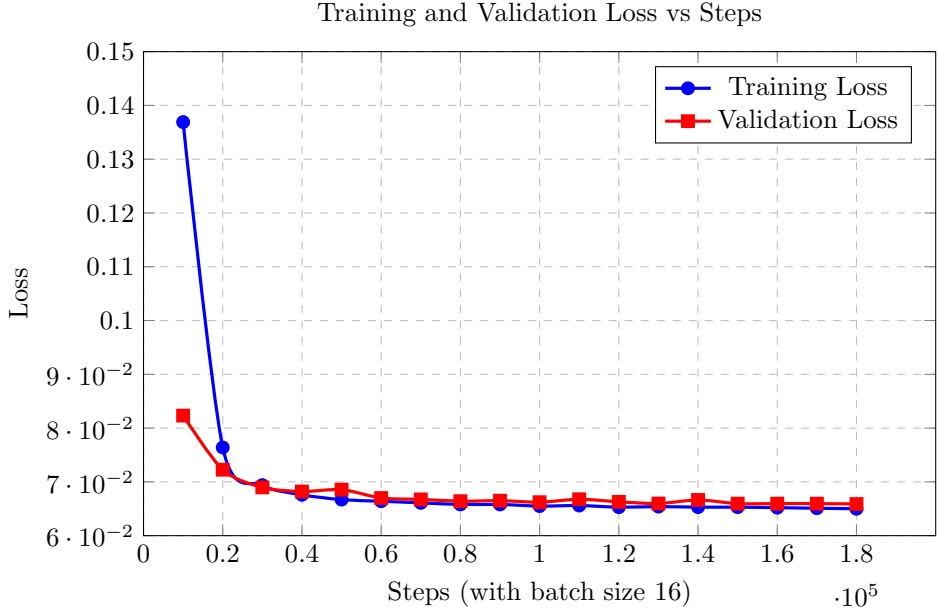
\begin{figure}[htbp]
    \centering
    \begin{tikzpicture}
    \begin{axis}[
        width=12cm, 
        height=8cm, 
        title={Training and Validation Loss vs Steps},
        xlabel={Steps (with batch size 16)},
        ylabel={Loss},
        xmin=0, xmax=200000,
        ymin=0.06, ymax=0.15,
        xtick={0,20000,40000,60000,80000,100000,120000,140000,160000,180000},
        ytick={0.06,0.07,0.08,0.09,0.10,0.11,0.12,0.13,0.14,0.15},
        legend pos=north east,
        ymajorgrids=true,
        xmajorgrids=true,
        grid style=dashed,
        smooth, 
    ]

    \addplot[
        color=blue,
        mark=*,
        line width=1.2pt,
    ]
    table[
        col sep=comma,
        x=Step,
        y=Training Loss,
    ]{loss_data.csv};
    \addlegendentry{Training Loss}

    \addplot[
        color=red,
        mark=square*,
        line width=1.2pt,
    ]
    table[
        col sep=comma,
        x=Step,
        y=Validation Loss,
    ]{loss_data.csv};
    \addlegendentry{Validation Loss}

    \end{axis}
    \end{tikzpicture}
\caption{Comparison of Training Loss and Validation Loss over Training Steps with batch size 16, for training Trajectory Language Model.}
    \label{fig:loss_plot}
\end{figure}

\subsection{Guided Reinforcement Learning} \label{ssec:grl}

In this subsection we will describe how \emph{AlgoPilot} uses the \emph{Trajectory Language Model} (\emph{TLM}) to guide the training of its reinforcement learning model described in Section \ref{ssec:rl}. As shown in Section \ref{ssec:rl}, although the model can learn to sort an array using \emph{Compare} and \emph{Swap} operations, the trajectory of such operations does not follow any algorithmic pattern. Therefore, we would like to add an additional reward (or penalty) to the reinforcement learning environment according to the \emph{TLM}. If an operation has high probability according to the \emph{TLM}, a relatively high reward should be given to agent of reinforcement learning. If an operation has low probability, a low or negative reward should be given.

Here we revisit the reward settings of the sorting environment. The reward of sorting success is 0.5, which uses a gamma of 0.99 (i.e., it barely decays over the steps). All other rewards have a gamma of 0.7, which decays quickly over several steps. Each step has a reward of -0.3, and each swap that moves a smaller element to the front has a reward of 1.0. 

Suppose $prob(a)$ is probability of an action $a$ according to the \emph{TLM}. We define the loss of action $a$ as

\begin{equation}
    loss(a) = -\ln{prob(a)}
\end{equation}

The additional reward is set to

\begin{equation*}
    TLMreward(a) = -0.02 \cdot loss(a)
\end{equation*}

\begin{figure}[ht]
\centering
\begin{tikzpicture}
  \begin{axis}[
      width=0.9\textwidth,
      height=8cm,
      xlabel={\#Episodes},
      ylabel={Success Rate},
      ymajorgrids=true,
      grid style=dashed,
      legend style={at={(0.97,0.72)}, anchor=north east},
      axis y line*=left,
      axis x line=bottom,
    ]
    \addplot[
      color=blue,
      mark=none,
      thick,
    ] coordinates {
      (1000,0.141) (2000,0.284) (3000,0.42) (4000,0.511) (5000,0.603)
      (6000,0.592) (7000,0.647) (8000,0.741) (9000,0.841) (10000,0.871)
      (11000,0.893) (12000,0.851) (13000,0.882) (14000,0.909) (15000,0.897)
      (16000,0.865) (17000,0.946) (18000,0.94) (19000,0.889) (20000,0.939)
      (21000,0.874) (22000,0.95) (23000,0.928) (24000,0.954) (25000,0.976)
      (26000,0.97) (27000,0.897) (28000,0.95) (29000,0.962) (30000,0.953)
      (31000,0.932) (32000,0.883) (33000,0.911) (34000,0.879) (35000,0.902)
      (36000,0.847) (37000,0.862) (38000,0.868) (39000,0.892) (40000,0.893)
      (41000,0.895) (42000,0.896) (43000,0.885) (44000,0.91) (45000,0.92)
      (46000,0.912) (47000,0.925) (48000,0.918) (49000,0.908) (50000,0.9)
      (51000,0.912) (52000,0.93) (53000,0.909) (54000,0.927) (55000,0.916)
      (56000,0.923) (57000,0.918) (58000,0.922) (59000,0.913) (60000,0.925)
      (61000,0.933) (62000,0.911) (63000,0.929) (64000,0.932) (65000,0.91)
      (66000,0.922) (67000,0.923) (68000,0.928) (69000,0.916) (70000,0.917)
      (71000,0.92) (72000,0.939) (73000,0.931) (74000,0.929) (75000,0.939)
      (76000,0.926) (77000,0.905) (78000,0.932) (79000,0.929) (80000,0.936)
      (81000,0.916) (82000,0.915) (83000,0.932) (84000,0.94) (85000,0.92)
      (86000,0.917) (87000,0.939) (88000,0.926) (89000,0.933) (90000,0.924)
      (91000,0.924) (92000,0.922) (93000,0.929) (94000,0.909) (95000,0.921)
      (96000,0.916) (97000,0.917) (98000,0.921) (99000,0.919)
    };
    \addlegendentry{Success Rate}
  \end{axis}

  \begin{axis}[
      width=0.9\textwidth,
      height=8cm,
      xlabel={\#Episodes},
      ylabel={\#Operations},
      ymajorgrids=false,
      axis y line*=right,
      axis x line=none,
      legend style={at={(0.97,0.6)}, anchor=north east},
    ]
    \addplot[
      color=red,
      mark=none,
      thick,
    ] coordinates {
      (1000,44.18) (2000,45.33) (3000,45.34) (4000,45.37) (5000,40.99)
      (6000,38.76) (7000,38.63) (8000,38.65) (9000,38) (10000,38.05)
      (11000,37.46) (12000,36.14) (13000,35.4) (14000,35.52) (15000,36.91)
      (16000,35.64) (17000,35.76) (18000,35.75) (19000,34.97) (20000,35.98)
      (21000,35.75) (22000,35.77) (23000,35.42) (24000,35.02) (25000,34.79)
      (26000,36.53) (27000,36.19) (28000,36.5) (29000,36.02) (30000,36.86)
      (31000,35.61) (32000,35.54) (33000,35.52) (34000,35.48) (35000,35.46)
      (36000,34.34) (37000,34.13) (38000,33.91) (39000,33.95) (40000,34.83)
      (41000,34.3) (42000,34.07) (43000,34.07) (44000,34.61) (45000,34.42)
      (46000,34.26) (47000,34.74) (48000,34.55) (49000,34.42) (50000,34.37)
      (51000,34.45) (52000,33.68) (53000,34) (54000,34.11) (55000,34.2)
      (56000,34.76) (57000,33.67) (58000,34.36) (59000,34.57) (60000,34.23)
      (61000,34.2) (62000,34.28) (63000,34.31) (64000,34.2) (65000,33.97)
      (66000,34.33) (67000,34.48) (68000,34.58) (69000,34.19) (70000,33.69)
      (71000,34.01) (72000,33.89) (73000,34.5) (74000,34.48) (75000,34.45)
      (76000,34.44) (77000,34.33) (78000,34.61) (79000,34.28) (80000,34.12)
      (81000,34.26) (82000,34.55) (83000,34.07) (84000,34.04) (85000,34.1)
      (86000,34.18) (87000,34.39) (88000,34.04) (89000,33.94) (90000,33.92)
      (91000,34.22) (92000,34.36) (93000,33.97) (94000,33.78) (95000,34.32)
      (96000,33.76) (97000,33.99) (98000,33.82) (99000,34.04)
    };
    \addlegendentry{\#Operations}
  \end{axis}
\end{tikzpicture}
\caption{Success Rate and \#Operations vs. \#Episodes for \emph{AlgoPilot} with Guided Reinforcement Learning}
\label{fig:grl}
\end{figure}
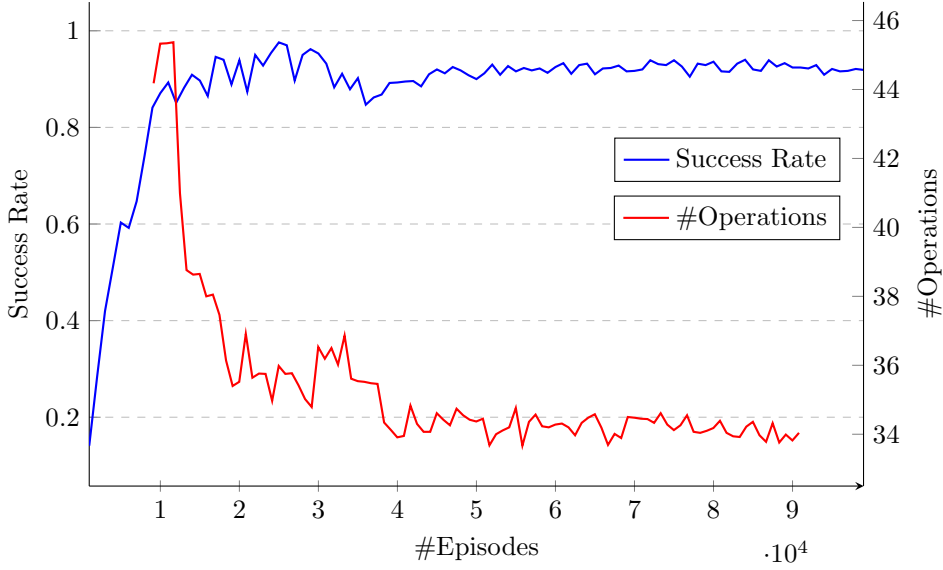

Let us first look at the success rate and number of operations for the above approach (for arrays of size 8). As shown in Figure \ref{fig:grl}, \emph{AlgoPilot} achieves about 95\% of success rate, with a reasonably low number of operations.

Here is an example trajectory after training for 10,000 episodes: \texttt{len8 Compare 0 1 more Swap Compare 1 2 more Swap Compare 2 3 more Swap Compare 3 4 more Swap Compare 4 5 more Swap Compare 5 6 more Swap Compare 6 7 more Swap Compare 1 2 less Compare 2 3 more Swap Compare 3 4 less Compare 4 5 more Swap Compare 5 6 more Swap Compare 0 1 more Swap Compare 2 3 less Compare 3 4 less Compare 4 5 more Swap Compare 1 2 more Swap Compare 2 3 more Swap Compare 3 4 more Swap Compare 4 5 more Swap Compare 1 2 less Compare 2 3 more Swap}. 

We feed the above trajectory into GPT-4o-mini, and ask it to generate a python function that is likely generate the above trajectory (with discrepancies allowed). The prompt is in Appendix \ref{appendixB}. The python function generated by GPT-4o-mini is shown below:

\begin{quote}

\begin{verbatim}
def generate_actions(a):
    actions = []
    n = len(a)

    # Compare and swap logic, similar to a bubble sort approach
    for i in range(n - 1):
        for j in range(n - i - 1):
            actions.append(f"Compare {j} {j + 1}")
            if a[j] > a[j + 1]:
                actions.append("more")
                # Swap elements if the left one is greater
                a[j], a[j + 1] = a[j + 1], a[j]
                actions.append("Swap")
            elif a[j] < a[j + 1]:
                actions.append("less")
            else:
                actions.append("equal")
    return actions

# Example usage
a = [8, 7, 6, 5, 4, 3, 2, 1]
actions = generate_actions(a)
for action in actions:
    print(action)
\end{verbatim}

\end{quote}

We can see that the above function is a perfect implementation of Bubble Sort. It is possible that GPT-4o-mini generates this algorithm because it knows Bubble Sort. But if we look at the trajectory, anybody with basic programming experiences would be able to figure out the algorithm, even she does not know Bubble Sort.

\begin{quote}

len8 Compare 0 1 more Swap Compare 1 2 more Swap Compare 2 3 more Swap Compare 3 4 more Swap Compare 4 5 more Swap Compare 5 6 more Swap Compare 6 7 more Swap \textbf{(missing Compare 0 1)} Compare 1 2 less Compare 2 3 more Swap Compare 3 4 less Compare 4 5 more Swap Compare 5 6 more Swap Compare 0 1 more Swap \textbf{(missing Compare 1 2)} Compare 2 3 less Compare 3 4 less Compare 4 5 more Swap \textbf{(missing Compare 0 1)} Compare 1 2 more Swap Compare 2 3 more Swap Compare 3 4 more Swap Compare 4 5 more Swap \textbf{(missing Compare 0 1)} Compare 1 2 less Compare 2 3 more Swap

\end{quote}

\begin{figure}[ht]
    \centering
    \begin{tikzpicture}
        \begin{axis}[
            width=0.8\textwidth,
            height=0.5\textwidth,
            xlabel={\#Episodes},
            ylabel={Discrepancies},
            grid=both,
            xtick={10000,20000,30000,40000,50000,60000,70000,80000,90000,100000},
            ymin=0,
            ytick={0,1,2,3,4,5,6},
            legend pos=north east
        ]
            \addplot[
                color=blue,
                mark=square*,
                thick
            ]
            coordinates {
                (10000,4) (20000,3) (30000,3) (40000,5) (50000,3)
                (60000,3) (70000,3) (80000,5) (90000,4) (100000,4)
            };
        \end{axis}
    \end{tikzpicture}
    \caption{Discrepancies vs. \#Episodes for trajectories generated by \emph{AlgoPilot} compared with Bubble Sort}
    \label{fig:discrepancy}
\end{figure}
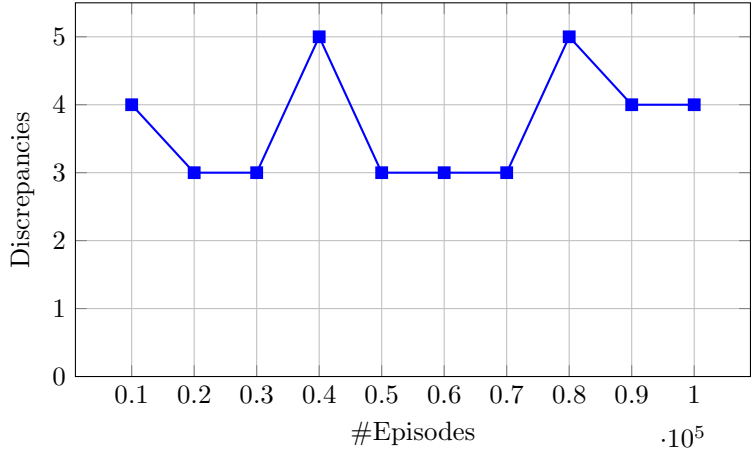

As shown above, the trajectory generated by our model (after 10,000 episodes) misses 4 comparisons, possibly because in most random arrays it is unnecessary to repeatedly compare the first two elements in each outer loop. We show the number of discrepancies for the trajectory generated by \emph{AlgoPilot}'s model trained for different numbers of episodes in Figure \ref{fig:discrepancy}. The number of discrepancies remains rather stable (between 3 and 5). We put each of the ten trajectories into GPT-4o-mini, and in each case a correct Bubble Sort algorithm is created.

\section{Discussions and Future Work}

Let us revisit the procedure of \emph{AlgoPilot} in creating an algorithm for using a double-loop to sort an array:

\begin{enumerate}
    \item \textbf{Random Function Generator}
    
        \par \noindent Use a random function generator to generate millions of random python function with a double-loop, and randomly add the two operations ``\emph{Compare}'' and ``\emph{Swap}''.
        
    \item \textbf{Trajectory Language Model (TLM)}
    
        \par \noindent Train a language model using the trajectories of the random functions, which can predict the probability of observing the next token after a prefix.
        
    \item \textbf{Reinforcement Learning}
    
        \par \noindent Use reinforcement learning to train a transformer model for sorting, using only \emph{Compare} and \emph{Swap} operations, with the environment only returning a simple feedback for each operation (\texttt{greater, equal, less} for \emph{Compare} and \texttt{sorted} for \emph{Swap}).
        
    \item \textbf{Guided Reinforcement Learning}
    
        \par \noindent Use the \emph{Trajectory Language Model (TLM)} to enhance the reinforcement learning process by adding an additional reward when the next operation has a high probability according to the \emph{TLM}. This guides the model to learn to generate a trajectory that can be produced by an algorithm.

    \item \textbf{Algorithm Creation}

        \par \noindent Use an LLM (such as GPT-4o-mini) to generate an algorithm or a python function based on the trajectory.
    
\end{enumerate}

We can see that the above procedure does not use any prior knowledge of a sorting algorithm, except that in Step 5 (\textbf{Algorithm Creation}) where the LLM is probably aware of all popular algorithms. Although it often looks obvious and trivial to create an algorithm based on a good trajectory, we still hope to get rid of any prior knowledge to achieve fully autonomy in algorithm creation.

Therefore, we will consider the creation of an algorithm based on a trajectory as our future work. We will use our random function generator to generate millions of functions and corresponding trajectories, and add random noise to the trajectories. Then we will train a sequnece-to-sequence model to predict the function based on the trajectory. It will be the last piece of puzzle to be added to the AI that can create an algorithm in a fully automated fashion.

\begin{appendices}

\section{Expected Numbers of \emph{Compare} and \emph{Swap} operations of Quicksort}\label{secA1}

\subsection{Model of QuickSort}

We consider the \textbf{classic} (textbook) QuickSort algorithm with the following characteristics:

\begin{enumerate}
    \item Pivot Selection: Choose one pivot uniformly at random from the current subarray.
    \item Partitioning Scheme: Utilize the \textbf{Lomuto} partition scheme:
    \begin{itemize}
        \item Compare each of the other \( n-1 \) elements to the pivot.
        \item Each time an element smaller than the pivot is found, increment a pointer and perform a swap.
        \item Finally, perform one additional swap to place the pivot in its correct position at the boundary.
    \end{itemize}
    \item Recursion: Recursively sort the subarrays to the left and right of the pivot.
\end{enumerate}

We aim to count:
\begin{itemize}
    \item \(\textbf{\#Comparisons}\): The number of times array elements are compared against the pivot.
    \item \(\textbf{\#Swaps}\): The number of times two elements in memory are exchanged.
\end{itemize}

Both counts are treated as random variables due to the random selection of pivots.

\subsection{Expected Number of Comparisons}

Let \( C(n) \) denote the (random) number of comparisons QuickSort makes on an array of size \( n \). The recurrence relation for the expected number of comparisons is:

\[
C(n) = (n - 1) + C(k) + C(n - k - 1),
\]

where:
\begin{itemize}
    \item \( (n - 1) \) is the number of comparisons in the current partition.
    \item \( k \) is the rank of the chosen pivot (\( 0 \leq k \leq n - 1 \)), selected uniformly at random.
    \item \( C(k) \) and \( C(n - k - 1) \) are the recursive calls on the resulting subarrays.
\end{itemize}

Taking expectations on both sides:

\[
\mathbb{E}[C(n)] = (n - 1) + \frac{1}{n} \sum_{k=0}^{n-1} \left( \mathbb{E}[C(k)] + \mathbb{E}[C(n - k - 1)] \right).
\]

Solving this recurrence leads to the closed-form expression involving harmonic numbers:

\[
\boxed{
\mathbb{E}[C(n)] = 2(n + 1) H_n - 2n
}
\]

where \( H_n = 1 + \frac{1}{2} + \cdots + \frac{1}{n} \) is the \( n \)-th harmonic number. For large \( n \), \( H_n \) approximates to \( \ln n + \gamma \) (Euler-Mascheroni constant, \( \gamma \approx 0.5772 \)), yielding:

\[
\mathbb{E}[C(n)] \sim 2n \ln n
\]

In base-2 logarithms, this can be expressed as:

\[
\mathbb{E}[C(n)] \approx 1.386\,n\,\log_2 n
\]

\subsection{Expected Number of Swaps}

Let \( S(n) \) denote the (random) number of swaps performed using the Lomuto partition scheme. The expected number of swaps in a single partition step is approximately:

\[
\frac{n - 1}{2} + 1 = \frac{n + 1}{2}
\]

Considering recursion, the recurrence relation is:

\[
S(n) = S(k) + S(n - k - 1) + \left( \text{swaps in current partition} \right),
\]

Taking expectations:

\[
\mathbb{E}[S(n)] = \frac{1}{n} \sum_{k=0}^{n-1} \left( \mathbb{E}[S(k)] + \mathbb{E}[S(n - k - 1)] \right) + \frac{n + 1}{2}
\]

Solving this recurrence yields:

\[
\boxed{
\mathbb{E}[S(n)] = (n + 1) H_n - 2n
}
\]

For large \( n \), this simplifies to:

\[
\mathbb{E}[S(n)] \sim n \ln n
\]

In base-2 logarithms:

\[
\mathbb{E}[S(n)] \approx 0.693\,n\,\log_2 n
\]

\subsection{Final Exact Formulas}

\begin{itemize}
    \item \textbf{Expected Number of Comparisons:}
    \[
    \boxed{
        \mathbb{E}[\text{\#Comparisons}] = 2(n + 1) H_n - 2n
    }
    \]
    
    \item \textbf{Expected Number of Swaps:}
    \[
    \boxed{
        \mathbb{E}[\text{\#Swaps}] = (n + 1) H_n - 2n
    }
    \]
\end{itemize}

Where \( H_n \) is the \( n \)-th harmonic number, defined as:

\[
H_n = 1 + \frac{1}{2} + \frac{1}{3} + \cdots + \frac{1}{n}
\]

As \( n \to \infty \), we have:

\[
H_n \sim \ln n + \gamma,
\]

where \( \gamma \) is Euler's constant. Therefore, the asymptotic behaviors are:

\[
\mathbb{E}[\text{\#Comparisons}] \sim 2n \ln n,
\]
\[
\mathbb{E}[\text{\#Swaps}] \sim n \ln n.
\]

\subsection{Caveats and Variations}

\begin{enumerate}
    \item \textbf{Partition Scheme:} The exact constants differ if using Hoare's partition scheme instead of Lomuto's. Hoare's scheme generally results in fewer swaps.
    
    \item \textbf{Pivot Selection Strategy:} Different pivot selection strategies, such as median-of-three, can alter the constants and lower-order terms in the expected counts, although the overall \( \Theta(n \log n) \) scaling remains unchanged.
    
    \item \textbf{Worst-Case Behavior:} Deterministic worst-case scenarios (e.g., always selecting the largest element as pivot) can lead to \( \Theta(n^2) \) comparisons and swaps. However, this is avoided in expectation with random pivot selection.
\end{enumerate}

\section{LLM prompt for algorithm generation from a trajectory} \label{appendixB}

Given a trajectory generated by our model in guided reinforcement learning, we use the following prompt to let GPT generate a python function that is likely to generate the trajectory (with small discrepancies allowed):

\texttt{Below is the sequence of actions of a python function that operates on an array of integers named "a", and its will receive feedbacks from an environment after it takes actions. Action "Compare i j" means the python function compares a[i] and a[j]. If a[i] is greater than a[j], the environment will return "more". If a[i] is smaller than a[j], it will return "less". Otherwise it will return "equal". Action "Swap" means the python function swaps a[i] and a[j]. Note this sequence of actions may contain noises, i.e., some actions may be missing and some others added. }

\texttt{Sequence of actions and feedbacks:}

\texttt{``len8 Compare 0 1 more Swap Compare 1 2 more Swap Compare 2 3 more Swap Compare 3 4 more Swap Compare 4 5 more Swap Compare 5 6 more Swap Compare 6 7 more Swap Compare 1 2 less Compare 2 3 more Swap Compare 3 4 less Compare 4 5 more Swap Compare 5 6 more Swap Compare 0 1 more Swap Compare 2 3 less Compare 3 4 less Compare 4 5 more Swap Compare 1 2 more Swap Compare 2 3 more Swap Compare 3 4 more Swap Compare 4 5 more Swap Compare 1 2 less Compare 2 3 more Swap''}

\texttt{Now, please write a python function that is likely to generate this sequence of actions. Your function should not generate the noises.}




\end{appendices}


\bibliography{sn-article}

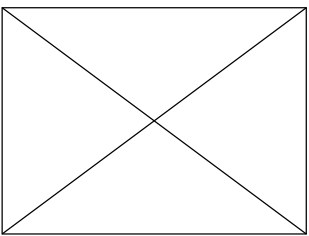
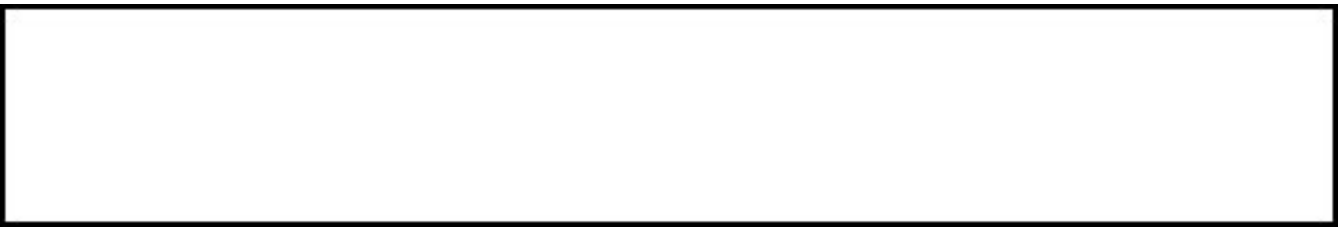
\end{document}